\documentclass[conference,onecolumn]{IEEEtran}
\IEEEoverridecommandlockouts
\usepackage{amsmath,amssymb,amsfonts}
\usepackage{algorithm}
\usepackage[pdftex]{graphicx}
\usepackage{textcomp}
\usepackage[utf8]{inputenc}
\usepackage{amsmath}
\usepackage{xcolor}
\usepackage{flushend}
\usepackage[hidelinks]{hyperref}
\usepackage{pdflscape}
\usepackage{tabularray}
\usepackage[T1]{fontenc}
\usepackage{cite}
\usepackage{fvextra}
\def\BibTeX{{\rm B\kern-.05em{\sc i\kern-.025em b}\kern-.08em
    T\kern-.1667em\lower.7ex\hbox{E}\kern-.125emX}}

\usepackage{multirow}
\usepackage{booktabs} 
\usepackage{algpseudocode}
\usepackage{etoolbox}
\makeatletter
\patchcmd{\@makecaption}
  {\scshape}
  {}
  {}
  {}
\makeatother

\definecolor{lightgray}{gray}{0.95}

\usepackage[most]{tcolorbox} 

\newtcolorbox{FVerbatim}{
  colback=lightgray, 
  colframe=black, 
  boxsep=0pt, 
  top=10pt, 
  bottom=10pt, 
  left=10pt, 
  right=10pt, 
  arc=0pt, 
  boxrule=1pt, 
  breakable, 
  before upper=\begingroup\alltt, 
  after upper=\endgroup 
}

\usepackage{graphicx}
\usepackage{times}
\usepackage{latexsym}
\usepackage{alltt,xcolor}
\usepackage{enumitem}

\usepackage{microtype}
\usepackage{booktabs}
\usepackage{multirow}
\usepackage{tikz}
\usepackage{graphicx}
\usepackage{fancyvrb}

\begin{document}

\newcommand\phuc[1]{\noindent{\color{blue} {\bf \fbox{Phuc}} {\it#1}}}
\newcommand\dat[1]{\noindent{\color{blue} {\bf \fbox{Dat}} {\it#1}}}
\newcommand\an[1]{\noindent{\color{blue} {\bf \fbox{An}} {\it#1}}}

\title{Rx Strategist: Prescription Verification using LLM Agents System}


\author{
Phuc Phan Van\IEEEauthorrefmark{1}, Dat Nguyen Minh\IEEEauthorrefmark{1}, An Dinh Ngoc\IEEEauthorrefmark{1}, Huy Phan Thanh\IEEEauthorrefmark{1} \\\
University Collaboration with Li Jinghong\IEEEauthorrefmark{2}, Dong Yicheng\IEEEauthorrefmark{2} \\
\IEEEauthorblockA{\IEEEauthorrefmark{1}\textit{FPT University, Ho Chi Minh Campus, Vietnam}}
\IEEEauthorblockA{\IEEEauthorrefmark{2}\textit{Japan Advanced Institute of Science and Technology, Ishikawa Campus, Japan}}
\{phucpvse170209, datnmse170570, andnse171386, huypt24\}@fpt.edu.vn, and  \{lijinghong-n, s2320035\}@jaist.ac.jp}
\maketitle

\maketitle

\begin{abstract}

To protect patient safety, modern pharmaceutical complexity demands strict prescription verification. We offer a new approach - Rx Strategist - that makes use of knowledge graphs and different search strategies to enhance the power of Large Language Models (LLMs) inside an agentic framework. This multifaceted technique allows for a multi-stage LLM pipeline and reliable information retrieval from a custom-built active ingredient database. Different facets of prescription verification, such as indication, dose, and possible drug interactions, are covered in each stage of the pipeline. We alleviate the drawbacks of monolithic LLM techniques by spreading reasoning over these stages, improving correctness and reliability while reducing memory demands. Our findings demonstrate that Rx Strategist surpasses many current LLMs, achieving performance comparable to that of a highly experienced clinical pharmacist. In the complicated world of modern medications, this combination of LLMs with organized knowledge and sophisticated search methods presents a viable avenue for reducing prescription errors and enhancing patient outcomes.

\end{abstract}


\section{Introduction}

Verifying prescriptions is an essential stage in the healthcare process that guarantees patient safety and the best possible results from treatments. However, studies have shown that a significant proportion of prescribed dosages are erroneous. For instance, A work analysis of medication errors in Vietnamese hospitals \cite{nguyen2015medication} found that roughly $40\%$ of doses prescribed to patients are incorrect in two urban public hospitals in Vietnam. Moreover, the availability of healthcare professionals, particularly in regions like Vietnam, is limited, exacerbating the issue. According to the Ministry of Health \footnote{\href{https://moh.gov.vn/documents/174521/1760801/3.++BC-BYT-+T\%E1\%BB\%95ng+k\%E1\%BA\%BFt+ng\%C3\%A0nh.pdf/481b5482-2b3c-4487-bfd6-20dd2601cb04}{Ministry of Health (MOH) 2023 Report}}, Vietnam has just $12.5$ doctors and $3.2$ graduate pharmacists for every $10,000$ people. This shortage of qualified personnel underscores the urgent need for advanced systems capable of automating and enhancing prescription verification without relying heavily on human resources.

Leveraging artificial intelligence (AI), particularly LLMs, as an assistant for healthcare providers offers a promising solution to mitigate prescription errors. AI-powered systems can rapidly analyze vast amounts of medical information, potentially identifying inconsistencies or potential issues with dosages, drug interactions, and contraindications.  However, current LLM systems face challenges in achieving reliable performance in this domain. Notably, the limited availability of real-world clinical data for AI model training raises concerns about their generalization to heterogeneous patient populations and varied clinical scenarios. Moreover, many LLMs rely on memorization rather than deep medical reasoning \cite{beyondimitationgame, reduce_hallucinate}, making them susceptible to hallucinations or incorrect answers when faced with unfamiliar or complex cases.

To address these challenges, we propose a novel LLM agent system designed specifically for prescription verification. Our system incorporates a sequence of specialized agents including 2 main tasks: indication verification and dose verification, each equipped with a unique combination of knowledge graphs, rule-based systems, and LLM components. This modular architecture enables a comprehensive analysis of each prescribed active ingredient, taking into account patient-specific information, the indicated condition, and established medical knowledge by combining the strengths of structured knowledge sources with the adaptability of LLMs. In addition, we introduce a specialized dataset focused on drug information and a novel methodology for knowledge retrieval. This dataset, combined with our retrieval approach, aims to enhance the system's robustness and overall performance. 

Chain-of-thought (CoT), introduced by \cite{chain_of_thought}, marks a pivotal advancement in improving the reasoning abilities of text-generation models. CoT enables models to generate intermediate steps in their thought processes, mirroring human problem-solving techniques. Research by \cite{zero-shot-learner} further demonstrated that certain prompts, like "Let's think step-by-step," can naturally induce CoT reasoning in LLMs. These breakthroughs have laid the groundwork for ongoing research aimed at enhancing the reasoning capabilities of LLMs.

To address the inherent knowledge limitations of LLMs, Retrieval Augmented Generation (RAG) has emerged as a prominent approach. RAG integrates LLMs with information retrieval systems, enabling them to access and utilize relevant external knowledge. This is typically achieved by embedding both the query and candidate documents in a shared vector space and then identifying the documents with the highest similarity to the query. Recent advancements in RAG have focused on improving retrieval accuracy. For instance, HyDE \cite{HyDE} enhances retrieval by generating a hypothetical answer to the query and then embedding it alongside the documents, allowing for a more nuanced comparison. Alternatively, Take a Step Back \cite{step_back} aims to identify the foundational knowledge documents most relevant to the query, potentially leading to more accurate and comprehensive responses.

Beyond traditional document retrieval, advanced RAG systems have increasingly turned to knowledge graphs (KGs) to improve retrieval accuracy and context understanding. KGs offer a structured representation of knowledge, capturing entities, relationships, and facts in a graph format. By integrating KG retrieval into RAG pipelines, researchers have been able to achieve several key advantages:
\begin{itemize}
    \item Structured Knowledge: KGs provide a structured representation of knowledge, enabling more precise retrieval and reasoning compared to unstructured text.
    \item Semantic Understanding: KGs capture the semantic relationships between entities, allowing RAG systems to better understand the meaning and context of queries.
    \item Multi-hop Reasoning: KGs can facilitate multi-hop reasoning, where the system navigates multiple relationships in the graph to answer complex questions.
    \item Explainability: KGs can provide a transparent explanation of the reasoning process by highlighting the relevant entities and relationships used to generate a response.
\end{itemize}

Recent works such as \cite{graph-rag} and \cite{KG-RAG} have demonstrated the effectiveness of incorporating KG retrieval into RAG. These systems leverage KG embeddings, graph traversal algorithms, and graph neural networks to identify relevant entities and subgraphs within the knowledge graph, providing the LLM with richer context and enabling more accurate and informative responses.

Besides optimizing in LLMs and RAG, many works integrate many LLMs together and also provide them tools for calling \cite{interactive_llm, llm_wizard, personal_llm_insight, potential_llm_agent} to boost the performance of LLMs. These works demonstrated that a multi-agent system, where individual agents specialize in different reasoning tasks and communicate via function calls, can outperform single-agent models on complex question-answering tasks. This architecture allows for a more modular and adaptable approach, where agents can leverage each other's expertise and collaboratively generate responses. As a result, LLMs can benefit from improved accuracy, a wider range of capabilities, and better handling of complex tasks that require multiple perspectives.


In summary, our main contributions are as follows: (1) we introduce a novel system flow for prescription verification, leveraging a multi-agent LLM architecture combined with knowledge graphs and rule-based systems; (2) we provide a specialized dataset focused on drug information and a novel knowledge graph-based retrieval methodology that significantly enhances the performance of our system; and (3) our system demonstrates exceptional performance across various metrics from models to human labels, outperforming current LLMs and achieving parity with highly experienced clinical pharmacists.

\begin{figure}
    \centering
    \includegraphics[width=0.6\linewidth]{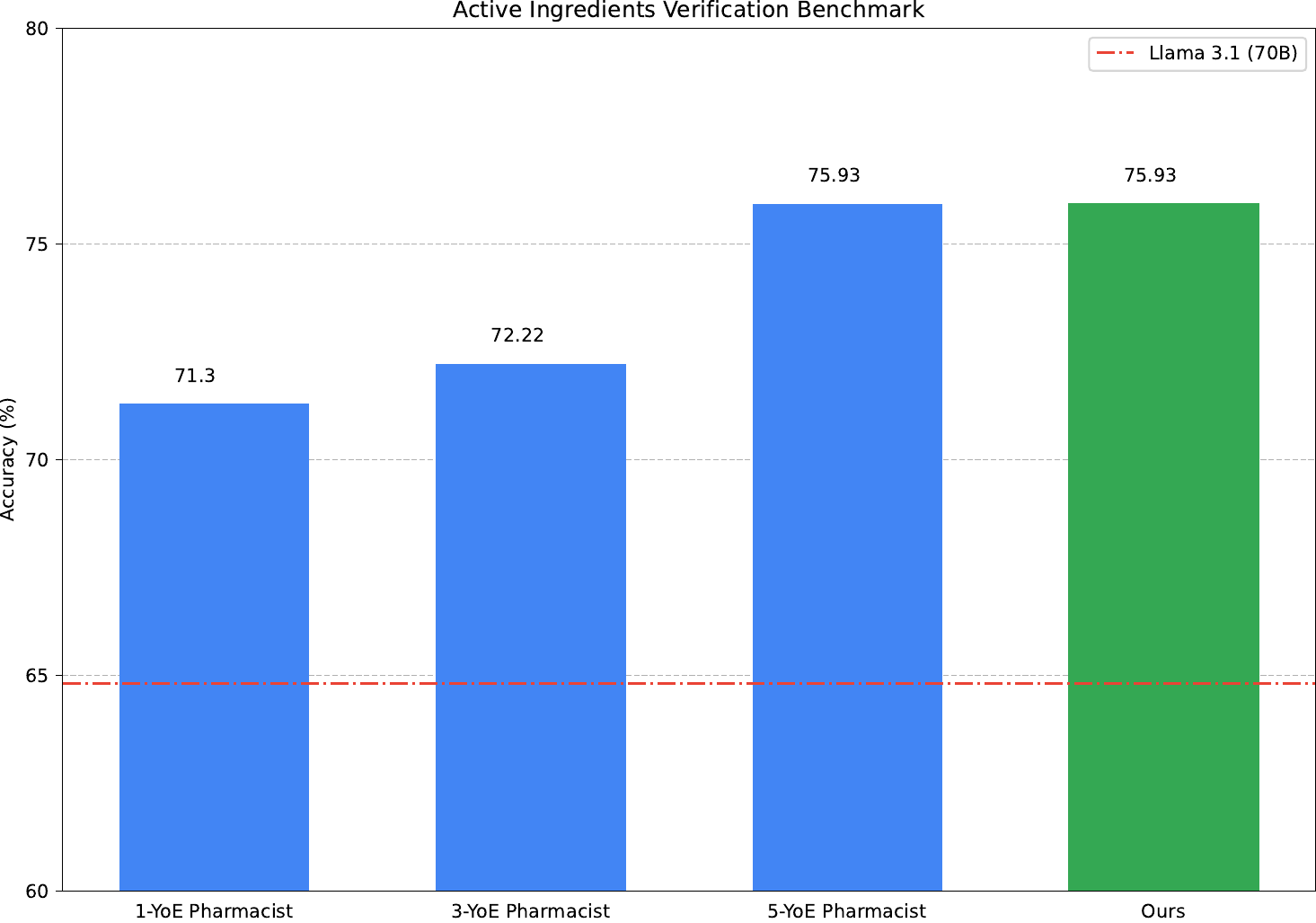}
    \caption{Benchmark of Drug Verification on our system against various levels of Clinical Pharmacists' Evaluation. The result shows that our system can perform at the level of senior pharmacists with five years of experience (YoE), while state-of-the-art LLMs (Llama 3.1 70B) fall short of even junior pharmacists' performance.}
    \label{fig:model-perf}
\end{figure}

\section{Dataset}

Current inference approaches for LLMs, such as CoT and ReAct \cite{yao2022react}, primarily rely on the model's response capabilities. However, these methods can be prone to hallucination and illogical reasoning. To mitigate these issues, we propose leveraging relevant reference materials, including drug and indication information, to guide LLM responses.

Instead of assessing specific drugs independently, we delve deeper into the \textbf{active ingredients}, which are specific compounds that make up a medication and are responsible for the therapeutic effects of it. For instance, Losartan is primarily used to treat hypertension and to protect the kidneys from damage due to diabetes, as well as helps relax blood vessels, making it easier for the heart to pump blood and reducing the risk of strokes and heart attacks \footnote{\href{https://www.drugs.com/losartan.html}{Losartan information}}.

\subsection{Data Collection} 


\begin{figure}
    \centering
    \includegraphics[width=\linewidth]{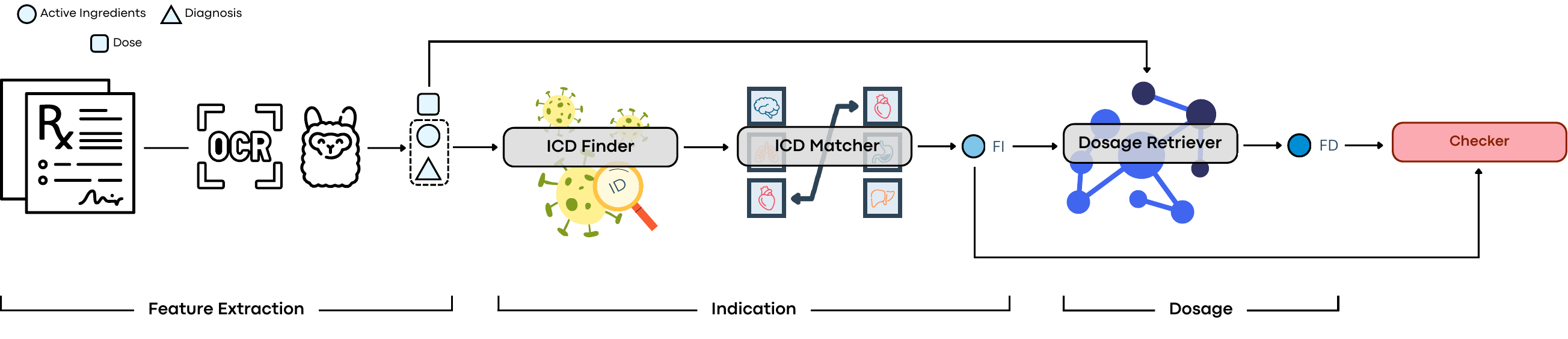}
    \caption{An overview of Rx Strategist. The process begins with extracting key information from the prescription, including the diagnosis, prescribed dosage, and active ingredients. This information is then passed to the indication verification module, which first identifies the ICD-10 code associated with the indicated condition and then cross-references it with the patient's diagnosis to ascertain whether the prescribed active ingredients are appropriate for treatment. Following this verification, the relevant active ingredients proceed to the dosage retriever module, which assesses whether the prescribed dosage falls within the recommended range for the patient's specific characteristics. Finally, the checker module consolidates the information from both the indication verification and dosage retriever stages, providing a comprehensive assessment and conclusion regarding the appropriateness of the prescription.}
    \label{fig:flow}
\end{figure}

\subsubsection{Drug Information Sources}
For this task, we collected highly accurate drug information from reliable sources like Drugs.com and Long Chau Pharmacy.
\begin{itemize}
    \item Drugs.com: Information on over 1700 active ingredients in AHFS DI Monographs format, including medication indication, administration, dosage, and adverse effects (if available).
    \item Long Chau Pharmacy: Vietnamese-language information on medicinal properties specific to Vietnam.
Data from both sources was stored and retrieved as HTML documents, with each document corresponding to a specific active ingredient. The data was then chunked according to headers and saved in Markdown format for human readability. For evaluation and querying purposes, the Markdown documents were further processed into a structured JSON format.
\end{itemize}


\subsubsection{Standardizing Indication Terminology}
To address inconsistencies in medical indication terminology, By using AI models like GPT 3.5 to generate ICD-10 codes based on various indication terms, we create a unified language for identifying and classifying diseases. This standardization allows healthcare professionals to quickly and easily identify the correct ICD-10 code, streamlining tasks such as medical billing, insurance claims processing, and research data analysis. In addition, accurate disease classification facilitated by standardized terminology can lead to improved diagnosis, treatment, and ultimately, patient outcomes.

\subsubsection{Drug Interaction Data}
We collected interaction data for 27 common active components, with detailed name in Appendix \ref{appendix:common_components_table}, from Drug.com interaction checker. The data include interaction level and detailed descriptions of their interactions with other components. By integrating this data, we aim to enhance the model's ability to identify and evaluate potential adverse effects, ensuring safer and more effective prescription recommendations.

\subsection{Expert-Approved Labels}

To ensure safe and accurate prescription verification, we leveraged the expertise of clinical pharmacists with varying levels of experience to establish a robust and reliable evaluation process. To assess the model's performance at different levels of human expertise, we enlisted three clinical pharmacists: a junior pharmacist with one year of experience, an intermediate pharmacist with three years of experience, and a senior pharmacist with five years of experience. Each pharmacist independently evaluated prescriptions, verifying the appropriateness of both the indication and dosage for each active ingredient. If an indication was deemed incorrect, the corresponding dosage was not evaluated. Our diligent human labeling process not only provided valuable insights but also optimized the evaluation of our automated verification model. 

A council of three highly experienced hospital pharmacists (with over five years of experience) provided the definitive assessment used as the gold standard for accuracy to which individual pharmacists' results were compared. This comprehensive evaluation strategy ensures that our system not only aligns with but surpasses industry standards, fostering trust in its ability to enhance patient safety and optimize medication management.

\subsection{Data Quality Control} 




For practical application and seamless integration into a database, the collected data underwent a rigorous preparation and quality control process. The initial Markdown format was converted into a structured JSON format for enhanced usability and compatibility.  This transformation involved several key steps:
\begin{itemize}
    \item Removal of extraneous characters: Non-essential symbols, including tabs (\textbackslash t) and daggers ($\dagger$), were eliminated. Newline characters (\textbackslash n) were retained to preserve content separation.
    \item Hyphen standardization: Long and short hyphens were unified to ensure consistency.
    \item Dosage unit conversion: Regular expressions were employed to convert dosage amounts expressed in micrograms (mcg) to the standard unit of milligrams (mg).
    \item Restructuring of JSON hierarchy: Redundant layers within the JSON structure were removed to streamline data access and manipulation.

\end{itemize}

\subsection{Data Statistics} 

\begin{figure}
    \centering
    \includegraphics[width=\linewidth]{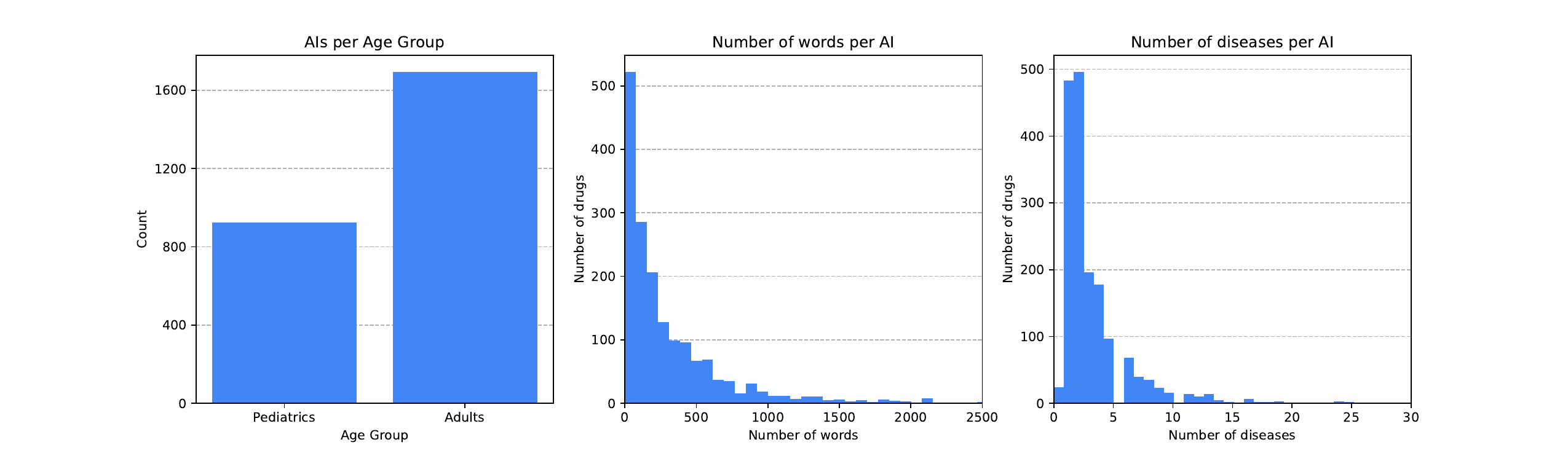}
    \caption{Some visualizations on the statistics of the active ingredient dataset. The left chart shows the discrepancy in adult and pediatric in quantity. The center chart displays the distribution in the number of words described in each drug. The right one indicates how many diseases each particular medicine can cure. AIs = Active Ingredients.}
    \label{fig:data-statistics}
\end{figure}

The dataset includes 1780 active ingredients, each containing information about compatible age groups alongside usage, as shown in Figure \ref{fig:data-statistics}.
\subsubsection{Age groups}
Close assession and filtering of dataset reveals that 1694 ($95\%$) of all available active ingredient can be used by adults, while only 923 ($52\%$) of them are prescribed towards pediatric patients. This imbalance of data can be understandable as the number of illnesses grows positively correlated with a person's age.
\subsubsection{Total information in each active ingredient}
In order to measure how much information is described in each medicine, we calculate the number of words available in the "description" portion of the data. The result shows that the number of words range from 0 up to 7697, with the majority of active ingredient descriptions fall under 1000 words. This indicates a large variance across our collection of active ingredient.
\subsubsection{Drug versatility}
The visualization also includes drug versatility measurement, which equals the total number of diseases one drug can cure, according to our dataset. On average, each drug can be indicated to cure 3 diseases, and few of them can reach up 40. 





\section{Methodology}

\textbf{Overview.} The Rx Strategist, as illustrated in Figure \ref{fig:flow}, is composed of three agents, an information extractor, and a checker. These components interact based on fit status, a key concept in the system, with the fit of indication (FI) and fit of dosage (FD) playing a crucial role in identifying the active ingredients and connecting the different modules. Specifically, a prescription is first processed by using optical character recognition (OCR) to extract relevant information, which is then converted into features by an LLM. \textbf{ICD Finder} is an LLM component identifies ICD-10 codes from the extracted active ingredients. \textbf{ICD Matcher} compares the identified ICD-10 codes with those in the patient's diagnosis, filtering the active ingredients to those that are suitable for the patient's condition as FI. \textbf{Dosage Retriever} determines the appropriate dosage of the active ingredient based on both FI and patient information. Finally, \textbf{Checker} utilizes both FI and FD to assess the overall validity of the prescription, providing explanations for its determination.

\subsection{Indication: Finding and Mapping}





To initiate the prescription verification process and provide the necessary input for the ICD Finder, we first extract relevant information from the prescription image. The image of a prescription is fed into an OCR model, returning a full-text format of the whole prescription. Subsequently, this body of text goes through a reformatting phase using GPT-4o-mini, turning it into a dictionary-like format (with information such as age group, indication, and dosage) for easy extraction in the latter phase.

\textbf{ICD Finder.} The ICD Finder receives a list of active ingredients in dictionary format. To address the challenge of inconsistent name representation, we employ a fuzzy matching algorithm to identify potential matches between the received active ingredients and the database entries. This fuzzy matching approach enhances the system's ability to recognize active ingredients with slight variations in naming conventions, thereby improving the accuracy of subsequent steps. Once potential matches are identified, the ICD Finder retrieves the corresponding usages for each active ingredient, indicating the diseases each active ingredient is designed to treat. Subsequently, the ICD Finder maps these diseases to their respective ICD-10 codes, establishing a set of ICD-10 codes associated with the prescription's intended therapeutic purposes.

In instances where information about a specific active ingredient is unavailable within our database, we leverage the capabilities of an LLM to generate potential ICD-10 codes. The LLM is trained on a vast corpus of medical literature and clinical data, enabling it to infer potential ICD-10 codes based on the active ingredient's name, chemical structure, and known therapeutic uses. However, it is important to note that LLM-generated ICD-10 codes should be treated with caution and may require additional validation before being used for clinical decision-making.

\textbf{ICD Matcher.} Leveraging the ICD-10 codes obtained by the ICD Finder, the ICD Matcher compares them against the ICD-10 codes derived from the patient's diagnosed conditions. The comparison is conducted at the category level of the ICD-10 code, which consists of an alphabetical letter followed by two numbers (e.g., A01, B15, C23). An active ingredient is labeled as "APPROPRIATE" if all ICD-10 categories associated with its usages are present in the patient's diagnosis codes, signifying that it aligns with the patient's medical needs. Conversely, if any ICD-10 category linked to an active ingredient's usage is absent from the patient's diagnosis codes, indicating a potential mismatch between the medication and the patient's condition, the active ingredient is classified as "INAPPROPRIATE".

\subsection{Dosage: Retriever with Knowledge Graph}

Due to the structured nature of our dataset and the possibility that LLMs might overlook subtle textual details \cite{beyondimitationgame}, we have developed a specialized text-to-graph processing approach. This method focuses on task-specific information extraction, improving the robustness of our model while minimizing the need for extensive fine-tuning.

We extracted dosage information for each disease within specific age groups (e.g., pediatric, adult) and administration routes (e.g., oral, intravenous) from our dataset. Given the absence of detailed patient histories in our prescription data, we established the initial dosage specified in the drug information as the recommended baseline dose. Any prescribed dosage deviating from this baseline was flagged for further scrutiny, as it might necessitate individualized adjustments based on the patient's specific circumstances and medical history. To structure this information, we constructed a knowledge graph with nodes representing drugs, diseases, and dosages, and edges denoting the relationships between them (see Appendix \ref{appendix:KG_structure} for details). The process of generating this knowledge graph is outlined in Algorithm \ref{alg:create_KG}. Specifically, the algorithm iterates through each active ingredient in the dataset, then for each age group associated with that ingredient, and finally for each disease that the ingredient is indicated for within that age group. For each disease, the language model is used to generate dosage nodes and relationships, which are then added to the knowledge graph. This structured representation facilitates efficient retrieval and reasoning about dosage information, enhancing the accuracy and effectiveness of our prescription verification system.

\begin{algorithm}[H]
\caption{Knowledge Graph Generation from Prescription Data}
\label{alg:knowledge_graph}
\begin{algorithmic}[1]
\Require{ActiveElementList, AgeGroupData, DiseaseData, LanguageModel} 
\Ensure{KnowledgeGraph}

\State KnowledgeGraph $\gets$ emptyGraph() \Comment{Initialize empty knowledge graph}

\For {each ActiveElement in ActiveElementList} 
    \For {each AgeGroup in AgeGroupData[ActiveElement]}
        \For {each Disease in DiseaseData[AgeGroup]}
            \State Output $\gets$ LanguageModel(Disease) \Comment{Let the model generate dosage nodes and relationships}
            \State add (ActiveElement, AgeGroup, Disease, Output) to KnowledgeGraph
        \EndFor
    \EndFor
\EndFor

\State \Return KnowledgeGraph
\end{algorithmic}
\label{alg:create_KG}
\end{algorithm}

\textbf{Dosage Retriever.} a knowledge graph-based system, efficiently retrieves appropriate dosages for verified active ingredients. It takes as input the active ingredients validated in the indication stage, along with patient-specific details like age group, age-specific factors (e.g., weight, kidney function), and the diagnosed condition the drug aims to treat. The system then navigates the knowledge graph, which is structured around relationships between drugs, diseases, and dosages. The retrieval process involves comparing the patient's information with the knowledge graph's nodes and edges. For instance, if the patient is a child diagnosed with a specific condition, the Dosage Retriever will traverse the graph to find the active ingredient node, follow the edge to the relevant disease node, and then identify the dosage node associated with the child age group. To enhance accuracy, a language model is integrated to standardize keywords and address inconsistencies in drug information representation. This ensures precise matching between the patient's data and the knowledge graph's information. If no exact match is found, the language model can suggest the closest available dosage information.

Algorithm \ref{alg:retrieve_KG} details the step-by-step process of dosage retrieval. It first identifies relevant diseases for the active ingredient and patient age group within the knowledge graph. If the diagnosed disease isn't directly linked, the language model assists in finding the closest match. The patient's age is then categorized into a specific age range, and potential dosages are retrieved based on the active ingredient, disease, and age range. If multiple dosage options exist, the system selects the most appropriate one based on the patient's specific age and any additional guidance from the language model. If no dosage information is available, the system indicates this lack of information.


\begin{algorithm}[H]
\caption{Prescription Dosage Retrieval (KG-based)}
\label{alg:dosage_retrieval}
\begin{algorithmic}[1]

\Require Knowledge Graph (KG), active ingredient (AE), Age Group (AG), Diagnosed Disease (D), Patient Age (PA), Language Model (LM)
\Ensure Recommended Dosage

\State RelevantDiseases $\gets$ KG.findDiseases(AE, AG) \Comment{Retrieve diseases treated by AE from KG}

\If {D not in RelevantDiseases}
    \State D $\gets$ LM.matchDisease(D, RelevantDiseases) \Comment{Use LM to match D to the closest disease in KG}
\EndIf
\State RelevantAgeRanges $\gets$ KG.findAgeRanges(PA)
\State AgeRange $\gets$ matchAgeRange(PA, RelevantAgeRanges)  \Comment{Categorize PA into age range (e.g., from 12 to 17 ages)}
\State DoseOptions $\gets$ KG.findDosages(AE, D, AgeRange) \Comment{Retrieve dosages based on AE, D, and AgeRange}

\If {DoseOptions is empty}
    \State \Return "No dosage information available" 
\Else
    \State \Return DoseOptions
\EndIf
\end{algorithmic}
\label{alg:retrieve_KG}
\end{algorithm}


\section{Experimental Settings}
\subsection{Benchmarks}
To evaluate the performance of our system, we curated a dataset of 20 real-world prescriptions sourced from Vietnam hospitals, the age group of patient is majoring as adults. To ensure patient privacy, all personally identifiable information, including names, addresses, and hospital details, was meticulously removed from the dataset.

\subsection{Baselines and Experimental settings}

Assessing our system’s capabilities reliably and comprehensively involved comparing its performance to a diverse set of benchmarks, including state-of-the-art language models (LLMs) and human experts. This multifaceted approach allowed for a robust evaluation across various metrics and perspectives. For the LLM baseline, we utilized both open-source (Qwen2 72B by Alibaba group \cite{yang2024qwen2}, LLama3.1 family models from 8 parameters to 405 parameters by Meta \cite{llama3.1}) and closed-source models (GPT4o-mini by OpenAI\footnote{\url{https://openai.com/index/gpt-4o-mini-advancing-cost-efficient-intelligence/}} and Claude 3.5 Sonnet by Anthropic\footnote{\url{https://www.anthropic.com/news/claude-3-5-sonnet}}). To evaluate human performance, we collected prescription evaluations from clinical pharmacists with 1 to 5 years of experience, establishing a real-world benchmark.

All LLM evaluations employed a CoT prompting method, with their specific prompts detailed in Appendix \ref{appendix:prompt}. Open-source models were initially configured with a temperature of 0, top-p of 0.7, and top-k of 50 for all models except LLama 3.1 8B. Additionally, We observed a tendency for LLama 3.1 8B to omit answers and duplicate words for certain prescription types. To address this, we adjusted the temperature for LLama 3.1 8B to 0.2 or 0.3 for prescription inference, and a value of 0.5 for interaction query summarization, encouraging a more exploratory approach and potentially leading to more comprehensive responses. Closed-source models were utilized with default settings provided by their respective developers.

Our system utilizes GPT4o-mini as the underlying language model due to its availability, ease of integration, and strong performance in preliminary tests.

\begin{table}[h!]
\caption{Table of Hyperparameter choices for Large Language Models as evaluation and comparison. The LLama3.1-8B model has a temperature value spanning from 0 to 0.5, for fine-tuned adjustments to output randomness. In contrast, the two closed-source models have their Top K parameter disabled by default}
\begin{center}
\begin{tabular}{lccc}
\toprule
Models & Temperature & Top k & Top p \\
\midrule
LLama3.1-8B & 0 - 0.5 & 50 & 0.7\\
\midrule
LLama3.1-70B & \multirow{3}{*}{0}  & \multirow{3}{*}{50} & \multirow{3}{*}{0.7} \\
Qwen2-72B &  & & \\
LLama3.1-405B & & & \\
\midrule
GPT4o-mini & 1.0 & - & 1.0 \\
Claude 3.5 Sonnet & 1.0 & - & 0.999 \\
\bottomrule
\end{tabular}

\end{center}

\label{tab:hyperparameters}
\end{table}

\subsection{Evaluation Metrics}
To assess the performance of our system, we prepare a variety of metrics including accuracy, precision, recall, F-0.5 score as our evaluation metrics. For each prescription, we compare the set of active ingredients predicted by our system against the corresponding gold standard set (i.e., the ground truth active ingredients).

\begin{table}[h!]
\caption{A comprehensive comparison of our Rx Strategist prescription verification system's performance against a diverse set of benchmarks. These include human experts (clinical pharmacists with 1, 3, and 5 years of experience (1Y, 3Ys, 5Ys)), open-source language models (LLama 3.1 family and Qwen 72B), and closed-source models (Claude 3.5 Sonnet and GPT4o-mini). Our results demonstrate that our system consistently outperformed these baselines across the majority of evaluated metrics.}
\begin{center}
\begin{tabular}{l|ccc|cccc|cc|c}
\toprule
\multirow{2}{*}{Metrics} & \multicolumn{3}{c|}{\textbf{Human}} & \multicolumn{4}{c|}{\textbf{Open-Source Models}} & \multicolumn{2}{c|}{\textbf{Closed-Source Models}} & \multirow{2}{*}{\textbf{Ours}} \\

 & 1Y & 3Ys & 5Ys & LLama3.1-8B & Llama3.1-70B & Qwen2-72B & LLama3.1-405B & Claude3.5-Sonnet & GPT4o-mini \\
\midrule

Accuracy & 71.30 & 72.22 & \textbf{75.93} & 56.48 & 64.81 & 68.52 & 74.07 & 72.22 & 70.37 & \textbf{75.93} \\

\midrule
Precision & 75.00 & 79.22 & 81.01 & 66.29 & 69.89 & 72.53 & 74.74 & 73.68 & 73.12 & \textbf{82.67} \\
\midrule
Recall & 88.00 & 81.33 & 85.33 & 96.72 & 86.67 & 88.00 & 94.67 & \textbf{100.00} & 90.67 & 82.67 \\
\midrule
F-$0.5$ Score & 77.28 & 79.63 & 81.84 & 70.74 & 72.71 & 75.17 & 78.02 & 77.78 & 76.06 & \textbf{82.67} \\
\bottomrule
\end{tabular}

\end{center}

\label{tab:main_table}
\end{table}

Basic metrics such as Accuracy, while reliable for most tasks, are not the best indicator of good performance for this task. More specifically, cases of inappropriate elements being flagged as appropriate are considered harmful to patients, as they create a false belief that their description is accurate when it is not. On the other hand, the opposite case of appropriate elements being flagged as inappropriate is less serious, as it only requires a close assession by the professionals. Therefore, we introduce more metrics to put emphasis on the harmful case.

The list of metrics we will use to assess performance is the following:
\begin{itemize}
    \item \textbf{Accuracy: }proportion of correct prediction over all samples.
    \item \textbf{Precision: }proportion of predicted appropriate elements that are correct.
    \item \textbf{Recall: } proportion of actual appropriate elements that are correctly predicted.
    \item \textbf{F-0.5 Score}: variant of F-$\beta$ Score with $\beta=0.5$, measuring weighted harmonic mean of precision and recall. Instead of the traditional F1-Score, which balances precision and recall, this variant weights more on precision, with the purpose of minimizing precision - inappropriate elements classified as appropriate.
\end{itemize}

\begin{figure}
    \centering
    \includegraphics[width=0.5\linewidth]{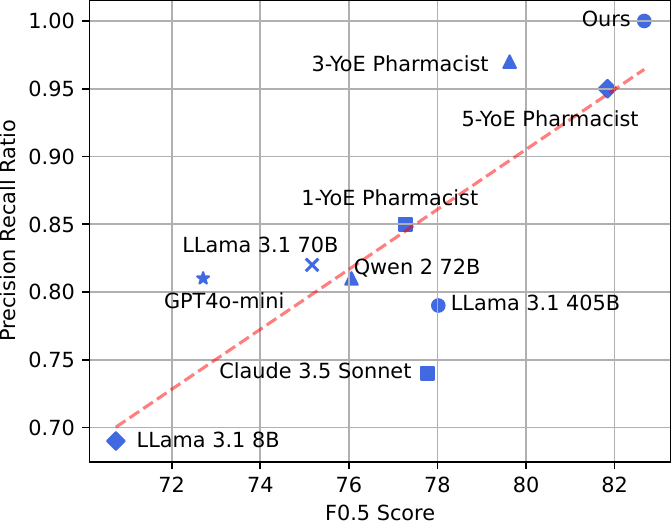}
    \caption{The relationship between the precision-recall ratio and the F0.5 score in prescription verification tasks. The prediction that strike a balance between precision and recall, thereby minimizing both false positives and false negatives, generally achieve higher F0.5 scores.}
    \label{fig:trend_ratio_fscore}
\end{figure}

\section{Results}

Table \ref{tab:main_table} presents a comprehensive comparison of our system's performance against human labels, open-source models, and closed models. The results unequivocally demonstrate that our system surpasses nearly all current LLMs, achieving a level of knowledge comparable to that of a clinical pharmacist with 5 years of experience in which you can see more in Figure \ref{fig:model-perf}.  In particular, our system outperforms GPT4o Mini by $5.56\%$ and Claude3.5 Sonnet by $3.71\%$ in terms of accuracy on closed models. On human labels, Rx Strategist outperforms clinical pharmacist with one year of experience by $4.63\%$ and by $3.71\%$ for clinical pharmacist with three years of experience, highlighting its superior capability in this domain.

\textbf{Interestingly, we observed a positive correlation between the precision-recall ratio and the F-0.5 score} in prescription verification tasks (Figure \ref{fig:trend_ratio_fscore}). This suggests that models prioritizing precision (minimizing false positives) while maintaining a reasonable recall (minimizing false negatives) tend to achieve higher overall performance, as indicated by the F-0.5 score. This finding highlights the importance of balancing precision and recall in this domain, where both accurate identification of valid prescriptions and avoidance of incorrect rejections are crucial. Our system, Rx Strategist, demonstrates this balance effectively, positioning it as a promising tool for enhancing prescription verification accuracy and ultimately improving patient safety.

Another noteworthy performance criterion is runtime, which is an important indication of how long our system can return its output. Table \ref{tab:inference_stat} clearly demonstrates the difference between the time taken by Rx-Strategist compared to other LLms, with additional information such as average time per token and total generated tokens for measuring conciseness.

\begin{table}[htbp]
\caption{The comparison of speed and token generated by Rx-Strategist against latest LLMs. While the fastest model (Llama 3.1 8B) is able to generate the output at twice the time as our solution, the majority of LLMs require much more tokens to output the whole answer. This indicates that our model is more efficient in token count while keeping the inference time low.}
\centering
\begin{tabular}{lccc}
\toprule
\textbf{Inference stat} & \textbf{Inference time (s)} & \textbf{Average time per token (ms)} & \textbf{Total generated tokens} \\
\midrule
LLama 3.1 8B & \textbf{5.29} & 803 & 15266 \\
LLama 3.1 70B & 10.89 & 794 & 15101 \\
Qwen 2 72B & 12.93 & \textbf{659} & 13192 \\
LLama 3.1 405B & 11.44 & 789 & 15789 \\
\textbf{Ours} & 10.50 & 705\footnotemark & \textbf{1223} \\
\bottomrule
\end{tabular}
\label{tab:inference_stat}
\end{table}\footnotetext{calculated for LLM only}


\section{Interaction Checking}

Given the complex nature of interaction between medical elements, we propose a new way of representing the relationship of medical properties by utilizing a Knowledge Graph. 
The triplets, which is used for graph construction, is extracted using the LLama3-8B model with prompting that targets domain-specific relationships, as detailed in the Appendix \ref{appendix:prompt_with_interaction}. Subsequently, the corresponding triplet embeddings are generated using the general text embedding model 
gte-Qwen2-1.5B-instruct \cite{li2023towards}. 

\begin{figure}
    \centering
    \includegraphics[width=0.9\linewidth]{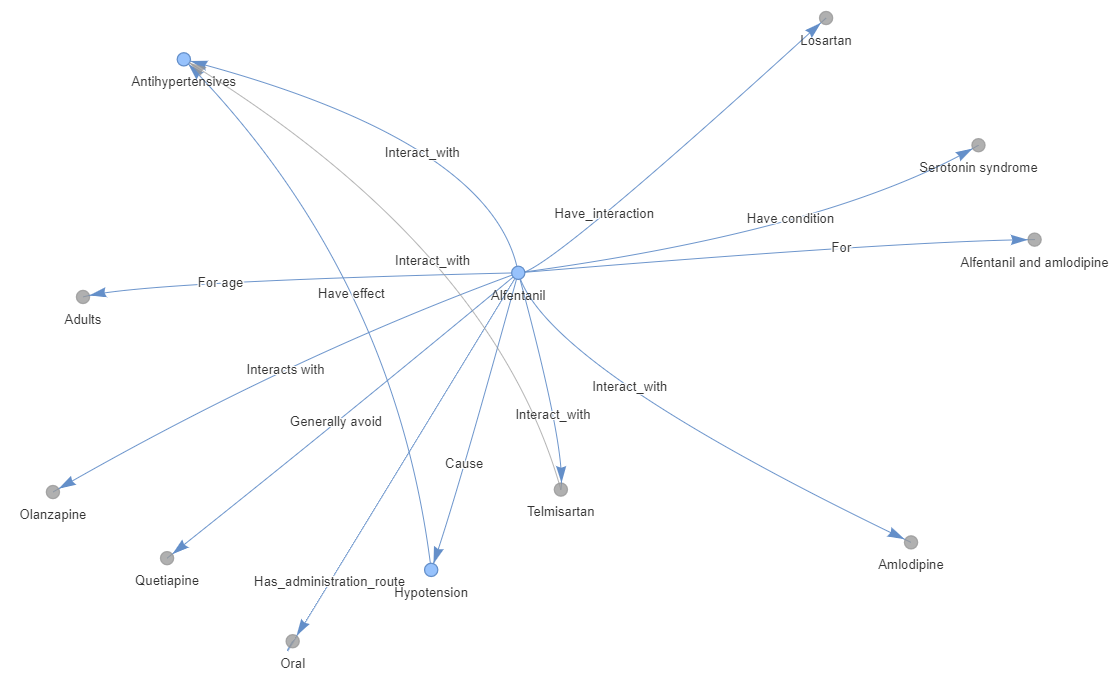}
    \caption{This is a sample of the element Afentanil and its related interaction. The center element branches to other elements connected by their corresponding relationship. The captured relationships are elements of interaction with Afentanil, the eligible age for usage such as Adults, and its adverse effects such as Hypotension.}
    \label{fig:KG-visualize}
\end{figure}

To validate the approach's performance against other experimental results and determine its effectiveness for our specific application, we have Knowledge Graph's impact on refining evaluation outcomes. The graph is combined with two models LLama3-8B \cite{llama3.0} and LLama3.1-8B \cite{llama3.1} respectively. Information on active components in prescriptions is retrieved from the graph using the cosine similarity score between the input query embedding and the precomputed triplet embeddings within the graph. The retrieved data is subsequently summarized using the same inference model to condense the information. We anticipate that this approach will enable the model to interpret the graph from its perspective. The summarized interactions are then utilized to aid the LLM in its prescription evaluation. The result shows that both LLama3-8B and LLama3.1-8B have improved their accuracy substantially, achieving an increase of $8.33\%$.

\begin{figure}
    \centering
    \includegraphics[width=0.6\linewidth]{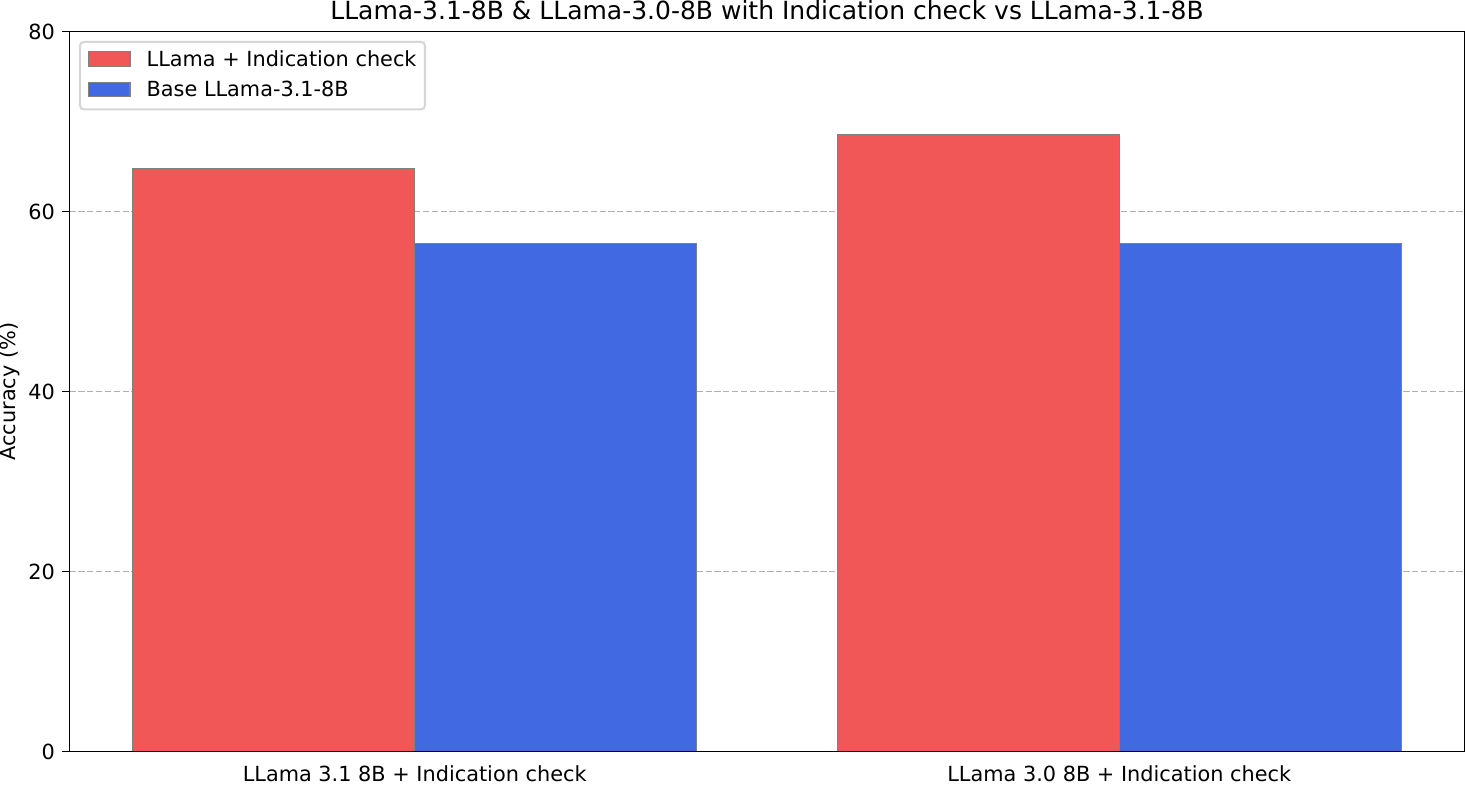}
    \caption{Benchmark of Drug Verification for base LLama3.1-8B compared to LLama3.1-8B and LLama3.0-8B combined with Interaction checking. The result shows that with the aid of Interaction Knowledge Graph, a small model can perform at the same level as larger LLMs.}
    \label{fig:llama3.1-8B-basemodel-compare}
\end{figure}



\section{Conclusion and Future work}

In conclusion, we present a novel, efficient approach to prescription verification, effectively combining a knowledge base with a well-instructed reasoning process. Our system's performance surpasses not only some senior clinical pharmacists but also state-of-the-art LLMs, demonstrating its potential for real-world implementation, particularly in resource-constrained hospital settings.

\textbf{Limitations and Future Works:} Several avenues for future enhancement exist. The current reliance on Vietnamese-language data necessitates further investigation into multilingual capabilities to ensure broader applicability. Additionally, refining the ICD-10 coding process, potentially through the development of a dedicated model, could mitigate the risk of hallucination and further improve accuracy. Expanding the knowledge base to incorporate diverse data sources, such as electronic health records and clinical guidelines, could enhance the system's understanding of complex medical scenarios, ultimately leading to more robust and reliable prescription verification.

\clearpage
\bibliographystyle{ieeetr}  
\bibliography{Bib_references}

\begin{thebibliography}{10}

\bibitem{nguyen2015medication}
H.-T. Nguyen, T.-D. Nguyen, E.~R. van~den Heuvel, F.~M. Haaijer-Ruskamp, and K.~Taxis, ``Medication errors in vietnamese hospitals: prevalence, potential outcome and associated factors,'' {\em PloS one}, vol.~10, no.~9, p.~e0138284, 2015.

\bibitem{beyondimitationgame}
A.~Srivastava, A.~Rastogi, A.~Rao, A.~A.~M. Shoeb, A.~Abid, A.~Fisch, A.~R. Brown, A.~Santoro, A.~Gupta, A.~Garriga-Alonso, {\em et~al.}, ``Beyond the imitation game: Quantifying and extrapolating the capabilities of language models,'' {\em arXiv preprint arXiv:2206.04615}, 2022.

\bibitem{reduce_hallucinate}
J.~Wei, Y.~Yao, J.-F. Ton, H.~Guo, A.~Estornell, and Y.~Liu, ``Measuring and reducing llm hallucination without gold-standard answers,'' 2024.

\bibitem{chain_of_thought}
J.~Wei, X.~Wang, D.~Schuurmans, M.~Bosma, B.~Ichter, F.~Xia, E.~Chi, Q.~Le, and D.~Zhou, ``Chain-of-thought prompting elicits reasoning in large language models,'' 2023.

\bibitem{zero-shot-learner}
T.~Kojima, S.~S. Gu, M.~Reid, Y.~Matsuo, and Y.~Iwasawa, ``Large language models are zero-shot reasoners,'' 2023.

\bibitem{HyDE}
L.~Gao, X.~Ma, J.~Lin, and J.~Callan, ``Precise zero-shot dense retrieval without relevance labels,'' 2022.

\bibitem{step_back}
H.~S. Zheng, S.~Mishra, X.~Chen, H.-T. Cheng, E.~H. Chi, Q.~V. Le, and D.~Zhou, ``Take a step back: Evoking reasoning via abstraction in large language models,'' 2024.

\bibitem{graph-rag}
D.~Edge, H.~Trinh, N.~Cheng, J.~Bradley, A.~Chao, A.~Mody, S.~Truitt, and J.~Larson, ``From local to global: A graph rag approach to query-focused summarization,'' 2024.

\bibitem{KG-RAG}
D.~Sanmartin, ``Kg-rag: Bridging the gap between knowledge and creativity,'' 2024.

\bibitem{interactive_llm}
Z.~Wang, G.~Zhang, K.~Yang, N.~Shi, W.~Zhou, S.~Hao, G.~Xiong, Y.~Li, M.~Y. Sim, X.~Chen, Q.~Zhu, Z.~Yang, A.~Nik, Q.~Liu, C.~Lin, S.~Wang, R.~Liu, W.~Chen, K.~Xu, D.~Liu, Y.~Guo, and J.~Fu, ``Interactive natural language processing,'' 2023.

\bibitem{llm_wizard}
K.~Yang, J.~Liu, J.~Wu, C.~Yang, Y.~R. Fung, S.~Li, Z.~Huang, X.~Cao, X.~Wang, Y.~Wang, H.~Ji, and C.~Zhai, ``If llm is the wizard, then code is the wand: A survey on how code empowers large language models to serve as intelligent agents,'' 2024.

\bibitem{personal_llm_insight}
Y.~Li, H.~Wen, W.~Wang, X.~Li, Y.~Yuan, G.~Liu, J.~Liu, W.~Xu, X.~Wang, Y.~Sun, R.~Kong, Y.~Wang, H.~Geng, J.~Luan, X.~Jin, Z.~Ye, G.~Xiong, F.~Zhang, X.~Li, M.~Xu, Z.~Li, P.~Li, Y.~Liu, Y.-Q. Zhang, and Y.~Liu, ``Personal llm agents: Insights and survey about the capability, efficiency and security,'' 2024.

\bibitem{potential_llm_agent}
Z.~Xi, W.~Chen, X.~Guo, W.~He, Y.~Ding, B.~Hong, M.~Zhang, J.~Wang, S.~Jin, E.~Zhou, R.~Zheng, X.~Fan, X.~Wang, L.~Xiong, Y.~Zhou, W.~Wang, C.~Jiang, Y.~Zou, X.~Liu, Z.~Yin, S.~Dou, R.~Weng, W.~Cheng, Q.~Zhang, W.~Qin, Y.~Zheng, X.~Qiu, X.~Huang, and T.~Gui, ``The rise and potential of large language model based agents: A survey,'' 2023.

\bibitem{yao2022react}
S.~Yao, J.~Zhao, D.~Yu, N.~Du, I.~Shafran, K.~Narasimhan, and Y.~Cao, ``React: Synergizing reasoning and acting in language models,'' {\em arXiv preprint arXiv:2210.03629}, 2022.

\bibitem{yang2024qwen2}
A.~Yang, B.~Yang, B.~Hui, B.~Zheng, B.~Yu, C.~Zhou, C.~Li, C.~Li, D.~Liu, F.~Huang, {\em et~al.}, ``Qwen2 technical report,'' {\em arXiv preprint arXiv:2407.10671}, 2024.

\bibitem{llama3.1}
H.~Touvron, T.~Culot, T.~Le~Scao, V.~Lam, S.~Edunov, J.~Wei, I.~Triantafillou, G.~Synnaeve, M.~Caron, P.~Martins, {\em et~al.}, ``The llama 3 herd of models,'' 2024.

\bibitem{li2023towards}
Z.~Li, X.~Zhang, Y.~Zhang, D.~Long, P.~Xie, and M.~Zhang, ``Towards general text embeddings with multi-stage contrastive learning,'' {\em arXiv preprint arXiv:2308.03281}, 2023.

\bibitem{llama3.0}
AI@Meta, ``Llama 3 model card,'' 2024.

\end{thebibliography}

\newpage
\appendix

\subsection{LLM Prompt for Evaluate}
\label{appendix:prompt}
\subsubsection{Base evaluation prompt}
{\scriptsize\begin{FVerbatim} 

You are a meticulous clinical pharmacist specializing in medication safety and appropriateness. Given a patient's
profile and prescription, your task is to thoroughly evaluate the prescription's suitability.
Pay close attention to the patient's age, medical conditions (especially kidney failure, liver disease, and
pregnancy), and any relevant allergies.
Follow these steps for each medication in the prescription:
1. \textbf{Assess Patient Profile:} Carefully review the patient's information to identify any potential risk factors
or contraindications.
2. \textbf{Verify Indication:} Determine if the medication or combination of medication is APPROPRIATE, INAPPROPRIATE, or
UNDERPRESCRIBED for the patient's diagnosed condition(s).
3. \textbf{Verify Dosage and Administration (If Appropriate):} If the medication is appropriate, confirm that the prescribed
dosage and administration instructions are safe and effective for the patient.
4. \textbf{Conclusion:} For each medication, provide a final assessment:
\hspace{10mm}* If APPROPRIATE, state "APPROPRIATE".
\hspace{10mm}* If INAPPROPRIATE, specify which aspect (e.g., dosage, active ingredient, interaction) is problematic and
provide a detailed explanation.
\textbf{Prescription:}
\end{FVerbatim}}

\subsubsection{Evaluation prompt with Interaction graph}
\label{appendix:prompt_with_interaction}
{\scriptsize\begin{FVerbatim}
You are a meticulous clinical pharmacist specializing in medication safety and appropriateness.
Given the reference materials bellow.
----------------------------------------
{Indication query summarization}
----------------------------------------
Your task is to thoroughly evaluate the prescription's suitability in english.
Pay close attention to the patient's age, medical conditions (especially kidney failure, liver disease, and
pregnancy), and any relevant allergies.
Follow these steps for each medication in the prescription:
1. \textbf{Assess Patient Profile:} Carefully review the patient's information to identify any potential risk factors or
contraindications.
2. \textbf{Verify Indication:} Determine if the medication or combination of medication is APPROPRIATE, INAPPROPRIATE,
or UNDERPRESCRIBED for the patient's diagnosed condition(s).
3. \textbf{Verify Dosage and Administration (If Appropriate):} If the medication is appropriate, confirm that the prescribed
dosage and administration instructions are safe and effective for the patient.
4. \textbf{Conclusion:} For each medication, provide a final assessment:
\hspace{10mm}* If APPROPRIATE, state "APPROPRIATE".
\hspace{10mm}* If INAPPROPRIATE, specify which aspect (e.g., dosage, active ingredient, interaction) is problematic and
provide a detailed explanation.
\textbf{Prescription:}
\end{FVerbatim}}

\subsection{Components table}
\label{appendix:common_components_table}
\begin{table}[h!]
\caption{Table of common drug components with collected Interaction Data.}
\centering
\begin{tabular}{lccc}
\toprule
Row id & Name\\
\midrule
1 & Allopurinol\\
2 & Amitriptyline\\
3 & Amlodipine\\
4 & Atorvastatin\\
5 & Bisoprolol\\
6 & Cefaclor\\
7 & Clopidogrel\\
8 & Dapagliflozin\\
9 & Dutasteride\\
10 & Empagliflozin\\
11 & Esomeprazole\\
12 & Gabapentin\\
13 & Isosorbide\\
14 & Ivabradine\\
15 & Losartan\\
16 & Meloxicam\\
17 & Metformin Hydrochloride\\
18 & Methylprednisolone\\
19 & Mirtazapine\\
20 & Nifedipine\\
21 & Olanzapine\\
22 & Pregabalin\\
23 & Quetiapine\\
24 & Rivaroxaban\\
25 & Rosuvastatin\\
26 & Spironolactone\\
27 & Telmisartan\\
\bottomrule
\end{tabular}

\end{table}

\newpage
\subsection{Data Representation}
\subsubsection{Raw JSON data Sample}
{\scriptsize\begin{FVerbatim}
\begin{Verbatim}[breaklines=True]
{"losartan": {
    "dosage": {
        "Pediatric Patients": {
            "Hypertension": {
                "Oral": "Children >6 years of age.. ."
            }
        },
        "Adults": {
            "Hypertension": {
                "Losartan Therapy": "Oral\nManufacturer recommends initial dosage of 50 mg..."
            },
            "Prevention of Cardiovascular Morbidity and Mortality": {
                "Oral": "Initially, 50 mg once daily..."
            },
            "Diabetic Nephropathy": {
                "Oral": "Initially, 50 mg once daily..."
            },
            "Heart Failure [off-label]": {
                "Oral": "Initially, 25-50 mg once daily ..."
            }
        }
    }
},
"esomeprazole": {
    "dosage": {
        "Pediatric Patients": {
            "GERD": {
                "GERD Without Erosive Esophagitis": "Oral\nChildren 1-11 years of age..."
            }
        },
        "Adults": {
            "GERD": {
                "GERD Without Erosive Esophagitis": "Oral\n20 mg once daily ..."
            },
            "Duodenal Ulcer": {
                "Helicobacter pylori Infection and Duodenal Ulcer": "Oral\nTriple therapy:..."
            },
            "NSAIA-associated Ulcers": {
                "Prevention of Gastric Ulcers": "Oral\n20 or 40 mg once daily;..."
            }
        }
    }
},
...
}
\end{Verbatim}
\end{FVerbatim}}

\newpage
\subsection{Prescription Sample}
\begin{figure}
    \centering
    \includegraphics[width=0.5\linewidth]{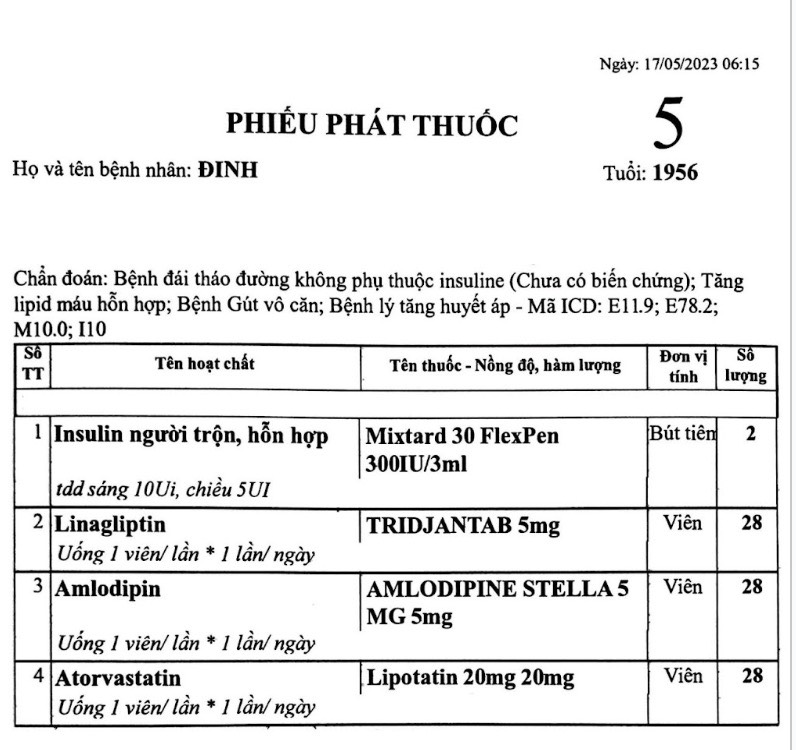}
    \caption{A sample of prescription which labeled as 5th, this prescription include some important information such as diagnosis, dose, and active ingredients.}
    \label{fig:prescription-samples}
\end{figure}

An example of a prescription named 24th is demonstrated in Figure \ref{fig:prescription-samples}.
\label{appendix:prescription}

\newpage
\subsection{Dosage Knowledge Graph Structure}
\label{appendix:KG_structure}
\subsubsection{Node Examples}
{\scriptsize\begin{FVerbatim}
\begin{Verbatim}[breaklines=True]
[
    {
        "id": 1,
        "name": "rosuvastatin",
        "type": "Drug"
    },
    {
        "id": 2,
        "name": "heterozygous familial hypercholesterolemia",
        "type": "Disease"
    },
    {
        "id": 3,
        "name": "5-10 mg once daily",
        "type": "Dosage"
    },
    ...
]
\end{Verbatim}
\end{FVerbatim}}
\subsubsection{Relationship Examples}
{\scriptsize\begin{FVerbatim}
\begin{Verbatim}[breaklines=True]
[
    {
        "start_id": 1,
        "end_id": 2,
        "type": "TREATS",
        "age_group": "pediatric"
    },
    {
        "start_id": 2,
        "end_id": 3,
        "type": "INITIAL_DOSAGE",
        "age_specific": "children 8 to <10 years of age",
        "administration": "oral"
    },
    {
        "start_id": 2,
        "end_id": 4,
        "type": "INITIAL_DOSAGE",
        "age_specific": "children and adolescents 10-17 years of age",
        "administration": "oral"
    },
    {
        "start_id": 1,
        "end_id": 5,
        "type": "TREATS",
        "age_group": "adults"
    },
    {
        "start_id": 5,
        "end_id": 6,
        "type": "INITIAL_DOSAGE",
        "administration": "oral"
    },
    {
        "start_id": 5,
        "end_id": 7,
        "type": "SPECIFIC_DOSAGE",
        "administration": "oral",
        "indication": "patients who have not achieved adequate response with the 20-mg daily dosage"
    },
    {
        "start_id": 1,
        "end_id": 8,
        "type": "TREATS",
        "age_group": "adults"
    },
    ...
]

\end{Verbatim}
\end{FVerbatim}}

\newpage
\subsection{Output Examples}
\subsubsection{LLama3.1 405B}
The output of the LLama 3.1 405B model on the 24th prescription (one of the prescription in the evaluation dataset). For more information about input, see at Appendix \ref{appendix:prescription}
{\scriptsize\begin{FVerbatim}
I'll evaluate each medication in the prescription based on the patient's profile.
\textbf{Patient Profile:}
- Age: 45 (born in 1978)
- Medical Conditions:
\hspace{10mm}- Chronic hepatitis B without hepatitis D coinfection
\hspace{10mm}- Non-insulin-dependent diabetes mellitus (without complications)
\hspace{10mm}- Mixed hyperlipidemia
\hspace{10mm}- Hypertension
- No allergies mentioned
\textbf{Medications}:
1. \textbf{Tenofovir (TDF) - Tefostad T300 300mg}
\hspace{10mm}- \textbf{Assess Patient Profile:} Patient has chronic hepatitis B, which Tenofovir is indicated for.
\hspace{10mm}- \textbf{Verify Indication:} APPROPRIATE for chronic hepatitis B.
\hspace{10mm}- \textbf{Verify Dosage and Administration:} The prescribed dose (300mg once daily) is standard for treating chronic
hepatitis B.
\hspace{10mm}- \textbf{Conclusion:} APPROPRIATE
2. \textbf{Atorvastatin - Lipotatin 20mg}
\hspace{10mm}- \textbf{Assess Patient Profile:} Patient has mixed hyperlipidemia.
   
\hspace{10mm}- \textbf{Verify Indication:} APPROPRIATE for hyperlipidemia.
   
\hspace{10mm}- \textbf{Verify Dosage and Administration:} The dose (20mg once daily) is within the recommended starting dose range
for adults.
   
\hspace{10mm}- \textbf{Conclusion:} APPROPRIATE

3. \textbf{Amlodipin + losartan - Troysar AM 5mg + 50mg}
   
\hspace{10mm}- \textbf{Assess Patient Profile:} Patient has hypertension.
   
\hspace{10mm}- \textbf{Verify Indication:} APPROPRIATE for hypertension management.
   
\hspace{10mm}- \textbf{Verify Dosage and Administration:} The combination and doses (5mg of Amlodipine and 50mg of Losartan once
daily) are within recommended ranges for hypertension management.
   
\hspace{10mm}- \textbf{Conclusion:} APPROPRIATE

4. \textbf{Metformin hydroclorid - Meglucon 1000 1000mg}
   
\hspace{10mm}- \textbf{Assess Patient Profile:} Patient has non-insulin-dependent diabetes mellitus.
   
\hspace{10mm}- \textbf{Verify Indication:} APPROPRIATE for type 2 diabetes.
   
\hspace{10mm}- \textbf{Verify Dosage and Administration:} The dose (1000mg twice daily) is a common starting dose for adults.
   
\hspace{10mm}- \textbf{Conclusion:} APPROPRIATE

5. \textbf{Linagliptin - TRIDJANTAB 5mg}
   
\hspace{10mm}- \textbf{Assess Patient Profile:} Patient has non-insulin-dependent diabetes mellitus.
   
\hspace{10mm}- \textbf{Verify Indication:} APPROPRIATE for type 2 diabetes.
   
\hspace{10mm}- \textbf{Verify Dosage and Administration:} The dose (5mg once daily) is the recommended dose for adults.
   
\hspace{10mm}- \textbf{Conclusion:} APPROPRIATE

\textbf{Overall Assessment:} All medications in the prescription are APPROPRIATE for the patient's conditions, with no
contraindications based on the provided information. However, it's essential to monitor kidney function, especially
with the use of Tenofovir and Metformin, and to assess the efficacy and potential side effects of all medications
regularly.

\end{FVerbatim}}

\newpage

\end{document}